# Witscript: A System for Generating Improvised Jokes in a Conversation

Joe Toplyn
Twenty Lane Media, LLC
P. O. Box 51
Rye, NY 10580 USA
joetoplyn@twentylanemedia.com

## Abstract

A chatbot is perceived as more humanlike and likeable if it includes some jokes in its output. But most existing joke generators were not designed to be integrated into chatbots. This paper presents Witscript, a novel joke generation system that can improvise original, contextually relevant jokes, such as humorous responses during a conversation. The system is based on joke writing algorithms created by an expert comedy writer. Witscript employs well-known tools of natural language processing to extract keywords from a topic sentence and, using wordplay, to link those keywords and related words to create a punch line. Then a pretrained neural network language model that has been fine-tuned on a dataset of TV show monologue jokes is used to complete the joke response by filling the gap between the topic sentence and the punch line. A method of internal scoring filters out jokes that don't meet a preset standard of quality. Human evaluators judged Witscript's responses to input sentences to be jokes more than 40% of the time. This is evidence that Witscript represents an important next step toward giving a chatbot a humanlike sense of humor.

## Introduction

For decades, people have imagined conversing with artificial entities as if they were human, and even deriving companionship from them. Recent advances in conversational AI have brought us closer to those social robots.

But to get even closer, a non-task-oriented conversational system—a social chatbot—needs to exhibit a sense of humor (Dybala, Ptaszynski, Rzepka, and Araki 2009a). The presence of humor improves a chatbot's performance and makes it seem more humanlike (Dybala, Ptaszynski, Rzepka, and Araki 2009b). Humor is especially important in non-task-oriented agents because their main purpose is to socialize with and entertain human users (Ptaszynski et al. 2010).

In addition to being able to recognize humor, an inviting conversational agent must also be able to generate it (Nijholt, Niculescu, Valitutti, and Banchs 2017). Even adding a simple pun-based joke generator to a chatbot can significantly improve its performance, quality, and likeability (Dybala, Ptaszynski, Higuchi, Rzepka, and Araki 2008).

However, existing systems for the computational generation of verbal humor have notable limitations. The virtual agents created by some large corporations only deliver canned jokes prewritten by humans (Nijholt et al. 2017). Other systems only generate self-contained jokes that aren't closely tied to a context (Amin and Burghardt 2020).

To be truly useful, a computational humor system needs to generate contextually integrated jokes about what's happening at the moment (Ritchie 2005). Such a system could improvise original and relevant joke responses to a user's utterances in a conversation, much as a witty human friend might.

One reason that existing joke generators fall short is that often they are not based on any overarching theory of linguistic humor (Amin and Burghardt 2020). Indeed, not many published theories of linguistic humor are detailed enough to serve as useful frameworks for devising computable algorithms (Ritchie 2009).

But a few theories do lead to algorithms, including the Surprise Theory of Laughter, proposed by a four-time Emmy-winning comedy writer in his book (Toplyn 2014). The Surprise Theory of Laughter says that we laugh when we're surprised that an incongruity turns out to be harmless.

The Surprise Theory of Laughter shares with the Benign Violation Theory (McGraw and Warren 2010) the idea that a necessary condition for laughter is that the incongruity be harmless.

The Surprise Theory of Laughter is also similar to the Two-Stage Model for the Appreciation of Jokes and Cartoons (Suls 1972), which is an incongruity-resolution model of humor. In the first stage of the Two-Stage Model, the audience encounters an incongruity—the punch line. In the second stage, the audience finds a cognitive rule that explains how the punch line follows from the preceding part of the joke, thus making sense of the incongruity. The Surprise Theory of Laughter differs from the Two-Stage Model in that the former stresses the importance of surprise: the audience must make sense of the incongruity suddenly, and therefore be surprised, if they are to laugh.

The Surprise Theory of Laughter provides the theoretical foundation for algorithms that Toplyn (2014) created to write the sort of monologue jokes delivered on comedy TV shows. Those algorithms specify that a monologue joke has three parts: the topic, the angle, and the punch line. The three parts appear in the joke in the following order:
1. The **topic** is the statement that the joke is based on.
2. The **angle** is a word sequence that smoothly bridges the gap between the topic and the punch line.
3. The **punch line** is the word or phrase that results in a laugh. It is a surprising incongruity that turns out to be harmless because it actually follows from the topic. This incongruity must appear at the end of the joke (Attardo 1994; Suls 1972).

Prewritten, canned jokes are structurally the same as jokes improvised during a conversation (Attardo 1994). Therefore, a system to improvise a conversational joke can be based on the above three-part structure of a canned monologue joke in the following way: A user's utterance to the system can be treated as the joke's topic. The improvised response generated by the system can be treated as the angle and punch line. So replying to the user's utterance with the system's response can complete a three-part joke that may amuse the user.

Toplyn (2014) also states that a three-part monologue joke can be written by taking certain steps in a particular order. I've distilled those steps into this **Basic Joke-Writing Algorithm**:

1. **Select a topic.** A good joke topic consists of one sentence that is likely to capture the attention of the audience for the joke.
2. **Select two topic handles.** The topic handles are the two words or phrases in the topic that are most responsible for capturing the audience's attention. The topic handles must capture the audience's attention because the audience must remember them in order to make sense of a punch line that is based on them. To remember information, we have to pay attention to it (Aly and Turk-Browne 2017).
3. **Generate associations of the two topic handles.** An association is something that the audience for the joke is likely to think of when they think about a particular subject. An association could be, for example, a person, place, thing, action, adjective, event, concept, or quotation.
4. **Create a punch line.** The punch line links an association of one topic handle to an association of the other topic handle in a surprising way.
5. **Generate an angle between the topic and punch line.** The angle is a sentence or phrase that connects the topic to the punch line in a natural-sounding way.

This paper presents Witscript, a system for generating improvised joke responses in conversations that is inspired by those algorithms created by a professional comedy writer (Toplyn 2014, 2020a). Human evaluators judged Witscript's improvisations to be jokes more than 40% of the time, compared with less than 20% for the output of an advanced conversational response generation model.

## Related Work

The Witscript system generates jokes that depend on wordplay, the clever manipulation of the sounds and meanings of words. Amin and Burghardt (2020) provide a comprehensive overview of existing systems for the computational generation of verbal humor, including systems based on wordplay and puns. But almost all of those wordplay systems generate self-contained puns instead of the contextually integrated puns that would be more useful in practical applications (Ritchie 2005).

As far as I know, only the PUNDA Simple system (Dybala et al. 2008) generates wordplay that is contextually integrated into a conversation as part of a response to a user's utterance. However, that system generates a joke by extracting a noun from a user's utterance, retrieving a sentence containing a punning word from a database, and responding to the user with a part of the retrieved sentence that starts with the punning word. By contrast, the Witscript system generates a wordplay punch line based on two words or word chunks extracted from the user's utterance. Witscript then places that punch line at the end of its joke response, as part of a sentence that in most cases is generated on the spot and not retrieved from a database.

Google's neural conversational model Meena (Adiwardana et al. 2020) has improvised at least one wordplay joke ("Horses go to Hayvard") in a conversation. But unlike Witscript, Meena has not been specifically designed to generate jokes. Therefore, the amount and timing of Meena's joke output can't be controlled.

## Description of the Witscript System

Toplyn (2014) presents six algorithms, called Punch Line Makers, for creating a punch line. The Basic Joke-Writing Algorithm underlies several of those Punch Line Makers, including the one that Witscript currently employs to generate a joke response to a user's utterance. Witscript uses Punch Line Maker #4: Find a play on words in the topic.

Wordplay can produce a humorous effect (Mihalcea and Strapparava 2005; Westbury and Hollis 2019). In terms of the Two-Stage Model (Suls 1972), Witscript uses wordplay to create an incongruous punch line that the user can make sense of by using a cognitive rule to connect it back to particular words in the user's utterance.

Because wordplay plays a central role in Witscript, first I'll discuss how the system calculates a wordplay score to measure the quality of the wordplay exhibited by two given words. Then I'll describe the rest of the system and how the wordplay score is used.

### Calculating the wordplay score

The Witscript system uses the pronunciation of words, i.e., their representation in phonemes, as given by the Carnegie Mellon University Pronouncing Dictionary (which is available at `http://www.speech.cs.cmu.edu/cgi-bin/cmudict`). The wordplay score for any pair of words is composed of these six subscores:

**Edit distance subscore**   This is the edit distance between either the pronunciations or, if a pronunciation is not in the dictionary, the spellings of the two words, as measured by the Levenshtein distance (Levenshtein 1966). Punning can include both phonetic and orthographical similarity (Valitutti, Doucet, Toivanen, and Toivonen 2015). The smaller the edit distance, the better the wordplay.

**Alliteration subscore**   This is 1 if the two words start with the same consonant phoneme. More points, which are constants, are added if more phonemes match. The greater this subscore, the better the wordplay. Alliteration and assonance tend to increase the funniness of a joke (Attardo, Attardo, Baltes, and Petray 1994; Mihalcea, Strapparava, and Pulman 2010).

**Assonance subscore**   This is a constant if the two words rhyme. If they don't, this subscore is the number of stressed vowels in one word that are repeated in the other word. The greater this subscore, the better the wordplay. The unexpected rhyming of two words that have different meanings can have a surprise effect (Attardo 1994), which, because the rhyming is harmless, results in funniness.

**Stop consonant subscore**   This is the total number of stop consonant phonemes (B, D, G, K, P, T) in both words. The greater this subscore, the better the wordplay. Stop consonants tend to make a joke funnier (Gultchin, Patterson, Baym, Swinger, and Kalai 2019; Toplyn 2014).

**Ending subscore**   This is 1 if the last phoneme of both words matches. More points, which are constants, are added if more phonemes match at the ends of the words. The final syllable is particularly important in wordplay (Attardo 1994). The greater this subscore, the better the wordplay.

**Syllable subscore**   This is 1 if the two words have the same number of syllables, which makes their pairing more improbable, more surprising, and therefore funnier.

To calculate the wordplay score for a pair of words, their six subscores are weighted and summed. The weights, along with the constants embedded in the subscores, were determined experimentally based on the quality of the system's output.

Now I'll describe Witscript as a whole and how it uses the wordplay score to create punch lines.

## Selecting two keywords from the topic

The process of generating a joke response begins when the Witscript system receives a sentence from a user, which it treats as the topic of a possible joke. From that sentence, the system extracts the nouns, noun phrases, and named entities using the Natural Language Toolkit (available from https://www.nltk.org/).

Any extracted nouns, noun phrases, and named entities that are on a list of stop words are excluded from consideration. The stop words include the most commonly used words plus other words that, in my expert opinion, are unlikely to be useful for generating a funny joke, words such as "official," "person," and "tonight."

The remaining nouns, noun phrases, and named entities that have been extracted become the candidates for topic keywords because the humor of human-created jokes tends to be based on nouns and noun phrases (Dybala, Ptaszynski, and Sayama 2011; West and Horvitz 2019).

The topic keywords will serve as the topic handles described above in the Basic Joke-Writing Algorithm. Therefore, the system selects as the topic keywords those two topic keyword candidates that are the most likely to capture the user's attention. The two topic keyword candidates that are the most likely to capture the user's attention are assumed to be the two candidates that are the least likely to appear together in the topic sentence.

To select the two topic keyword candidates least likely to appear together in the topic sentence, the system uses word embeddings created by Word2Vec (Mikolov, Chen, Corrado, and Dean 2013). Using Word2Vec word embeddings for this purpose seems reasonable because if those embeddings indicate that two words are unlikely to appear together in a large, general, training corpus, then those two words are also unlikely to appear together in a topic sentence received from a generic user during open-domain chat.

The Gensim library (Rehurek and Sojka 2010) is used to load a pretrained Google Word2Vec model (`GoogleNews-vectors-negative300.bin`) which was trained on about 100 billion words of Google News data. Then the "similarity" function of the Gensim implementation of Word2Vec is used to select the two topic keyword candidates that have the least cosine similarity. Those candidates become the topic keywords. For example, when the user inputs, "I just read that some flower that smells like a corpse is about to bloom," the system selects as the topic keywords "flower" and "corpse."

Next the system tries to link those two selected topic keywords by means of wordplay to create an incongruous punch line. Three types of wordplay punch lines are attempted: juxtaposition, substitution, and portmanteau.

## Creating a juxtaposition punch line

This type of punch line consists of two words right next to each other that, incongruously, exhibit wordplay. Juxtaposing two words can yield an incongruity that produces humor (Gultchin et al. 2019).

To create a juxtaposition punch line, the system starts by listing the top fifty words that are most closely associated with each of the two topic keywords; the number fifty was determined experimentally based on the quality of the system's output. To select those associated words, which I call associations, the system uses the "most similar" function of the Gensim implementation of the Google Word2Vec model trained on Google News data.

Next the system pairs the first topic keyword and each of its fifty associated words with the second topic keyword and each of its fifty associated words. The system selects the pair of words with the best wordplay score, one word from each list, to be the juxtaposition punch line. For the

example above, Witscript derives the juxtaposition punch line "garden carcass" from "flower" and "corpse."

### Creating a substitution punch line

This type of punch line consists of a multi-word chunk into which a new word has been substituted, incongruously, for a word with which it has wordplay. This word substitution can produce a humorous effect (Binsted and Ritchie 1994; Valitutti et al. 2015).

The system creates a substitution punch line by pairing each topic keyword with each of the words in every multi-word chunk in the top fifty "most similar" associations of the other topic keyword. The system selects the pairing with the best wordplay score. Then it substitutes one word in that pair for the other in the relevant multi-word chunk to create the substitution punch line.

For example, when the user inputs, "People are trying to summon a Mexican demon by getting him to spin a pencil," Witscript derives the substitution punch line "Puerto Demon" from "Puerto Rican," which it associates with "Mexican," and "demon."

### Creating a portmanteau punch line

This type of punch line entails the incongruous blending of two words into a portmanteau. The syllable substitution involved can result in humor (Binsted and Ritchie 1994).

The system pairs each topic keyword and its top fifty associations with the other topic keyword and finds a pair in which one word has a pronunciation that is similar to, but not identical to, the pronunciation of the beginning of the other word.

Then the system uses the Pyphen module (available from `https://pyphen.org/`) to divide the longer word into syllables, allowing the shorter word to be substituted for the equivalent number of syllables at the beginning of the longer word. This creates the portmanteau punch line.

For example, when the user inputs, "Researchers at Johns Hopkins have discovered a virus that causes stupidity," Witscript derives the portmanteau punch line "flupidity" from "flu," which it associates with "virus," and "stupidity."

### Selecting the best punch line

The system attempts to generate one punch line candidate of each of the above three types. Each punch line candidate that is generated has a wordplay score. The wordplay score of a juxtaposition punch line candidate is the wordplay score that was calculated for the two words selected to be the juxtaposition punch line. The wordplay score of a substitution punch line candidate is the wordplay score of the words that were substituted for one another to create the substitution punch line. The wordplay score of a portmanteau punch line candidate is the wordplay score of the word and syllables that were substituted for one another to create the portmanteau punch line.

Whichever punch line candidate has the best wordplay score is selected for inclusion in the system's joke response. This is because the best wordplay score is a proxy for the biggest incongruity. The punch line that embodies the biggest incongruity is the most surprising. And the punch line that is the most surprising is most likely the funniest (Suls 1972; Toplyn 2014).

The system filters out any punch line candidate that has a wordplay score worse than a preset, empirically determined threshold. If the system hasn't generated any punch line candidate with a wordplay score better than or equal to the threshold, then it doesn't output any joke response.

### Adding an angle to the selected punch line

After the system selects the best punch line, it adds an angle, which is text intended to smoothly connect the punch line to the user's input sentence.

To generate text to fill that gap, I used the language model BERT (Devlin, Chang, Lee, and Toutanova 2019) and the resources of Hugging Face, starting with their BERT, large, uncased model (available from `https://huggingface.co/bert-large-uncased`). That model had been pretrained for masked language modeling and next sentence prediction on BookCorpus, a dataset consisting of 11,038 unpublished books, and English Wikipedia (excluding lists, tables, and headers).

To fine-tune that pretrained BERT model, I used a dataset of late-night TV comedy show monologue jokes. To create that dataset, I first scraped 43,145 jokes from the archives available at these three websites:
`https://www.newsmax.com/jokes/archive/`
`https://github.com/brendansudol/conan-jokes-data`
`http://www.101funjokes.com/political_jokes.htm`

Next I prepared the dataset by taking the following steps: 1) I removed any duplicate jokes; 2) I added a topic to any joke that did not explicitly include one because, for example, the joke was a follow-up joke that implicitly assumed the topic of the previous joke; 3) I removed any extraneous words from the beginning and end of each joke like, for example, lead-in or concluding text that was only tangentially related to the actual joke; 4) I removed any joke that wasn't at least 17 tokens long because inspecting the dataset revealed that jokes shorter than that tended to be formed unclearly, lacking a distinct topic, angle, or punch line; 5) I removed any joke that was longer than 49 tokens because inspecting the dataset revealed that those longer jokes tended to be so wordy that the individual parts of the actual joke were hard for me to discern; 6) I manually annotated the remaining 36,781 jokes by splitting each of them into two segments—the topic and the rest of the joke—thus formatting them to use as input for fine-tuning the BERT model.

I used the prepared dataset of jokes to fine-tune the pretrained BERT model for one epoch and also for two epochs. Then I tested those two fine-tuned models, and the pretrained BERT model without any fine-tuning, using as input the topics and punch lines of 48 monologue-style jokes I had written that were not part of the fine-tuning dataset. The BERT model that had been fine-tuned for one epoch generated the best angle, in my judgment, considerably more often than the other two models did. So the model fine-tuned for one epoch became the joke-tuned BERT model that the Witscript system uses.

Here's how the Witscript system generates an angle using the joke-tuned BERT model. The BERT model is given the selected punch line, to which a [MASK] token has been prepended, together with the user's input sentence. BERT predicts a token to fill the mask position and then prepends that token to the punch line. Predicting tokens backward from the punch line in this way seems empirically to produce more natural-sounding output than predicting forward from the topic. BERT continues prepending [MASK] tokens and predicting tokens until it generates a stop condition, such as a punctuation mark or a repeated token. Whatever text BERT generated before the stop condition becomes the angle, which is prepended to the previously selected punch line to form Witscript's basic joke response.

Occasionally the very first token that BERT predicts is a punctuation mark, which indicates that BERT will be generating gibberish that won't be useable as an angle. In that case, the system selects an angle template at random from a list I wrote and inserts the selected punch line into it to form Witscript's basic joke response.

## Finalizing the joke response

The Witscript system prepends to the basic joke response a filler word randomly selected from another list I wrote. Filler words such as "um" and "like" make the response sound more humanlike (Duvall, Robbins, Graham, and Divett 2014).

The system also repeats the two topic keywords at the beginning of its response, to help the user find the connections between the user's input and the punch line; the system sometimes makes connections that are a bit obscure, and if a joke is too hard to process, it won't be funny (Attardo et al. 1994).

Then the system outputs its final joke response to the user. Here's how the system responds to the example user inputs above:

**Input**: "I just read that some flower that smells like a corpse is about to bloom."
**Witscript**: "Flower corpse? Heh, so now it smells like a garden carcass."

**Input**: "People are trying to summon a Mexican demon by getting him to spin a pencil."
**Witscript**: "Mexican demon? Mmm-hmm, or a Puerto Demon."

**Input**: "Researchers at Johns Hopkins have discovered a virus that causes stupidity."
**Witscript**: "Virus stupidity? Um, and not because of flupidity."

As an illustration of how Witscript produces its output, consider that last example:
- "Virus stupidity?" and the filler word "Um" were added when finalizing the joke response.
- The angle "and not because of" was generated by the joke-tuned BERT model.
- The punch line "flupidity" was created as described above in the section "Creating a portmanteau punch line" and selected as the best punch line.

## System Evaluation

To evaluate the Witscript system fairly, I wanted to compare it to a strong baseline. It seemed to me that a strong baseline would not be a model that generates random responses to input sentences, but instead a model that generates joke responses that attend to the input sentences.

But to my knowledge no such input-attentive joke generators exist that work in English; the PUNDA Simple system (Dybala et al. 2008) generates joke responses in Japanese. So, without access to a demo of Google's Meena (Adiwardana et al. 2020), I decided to use Microsoft's neural conversational response generation model DialoGPT (Zhang et al. 2020) as a baseline model. Although DialoGPT was not designed to generate joke responses, at least it can generate a response that attends to an input sentence.

To evaluate Witscript, I selected 20 monologue-type jokes that professional comedy writer Joe Toplyn had written and posted on his Twitter account @JoeToplyn. None of those 20 selected jokes had been used to fine-tune BERT. The topic sentence of each selected Twitter joke became an input for testing the Witscript and DialoGPT systems. The part of each selected Twitter joke that came after its topic sentence became the gold-standard, human response to that input. That is, each selected Twitter joke provided an input for testing and also the human response to that input.

All of the 20 Twitter jokes selected to evaluate the system met the following criteria:
- Their punch lines, like Witscript's, featured wordplay. That requirement minimized the effect of any bias introduced by human evaluators who don't like wordplay jokes.
- Their topic sentences appeared to have at least two nouns, noun phrases, or named entities that were in the vocabulary of the Google Word2Vec model. That made it more likely that Witscript would output responses.
- Their topic sentences didn't include any named entities for which the Google Word2Vec model was likely to yield stale associations. That way, Witscript wouldn't be penalized for having to rely on a model trained on an old, static, news dataset.

I input the topic sentences of the 20 selected Twitter jokes into Witscript but only received responses from Witscript for 13 of those topic sentences. For the topic sentences of the other 7 selected Twitter jokes, Witscript was apparently unable to generate any punch line candidate that had a wordplay score better than or equal to its internal threshold.

The 13 topic sentences for which Witscript did output responses were then used as input to obtain 13 responses from DialoGPT. To obtain responses from DialoGPT, I used Hugging Face's implementation of the model DialoGPT-large, which had been trained on dialogue from Reddit discussion threads (available from `https://huggingface.co/Microsoft/DialoGPT-large`). I started a new chat before inputting each topic sentence to DialoGPT, so as to eliminate any influence of dialogue history on the responses of DialoGPT.

I hired workers via Amazon Mechanical Turk to evaluate the responses generated by Witscript, DialoGPT, and the human for each of the 13 input sentences. The only qualifications that I specified for the workers were that they had to be located in the United States and have a Human Intelligence Task (HIT) Approval Rate greater than or equal to 95%. Therefore, nothing else is known about the evaluators.

The evaluators were not told how any of the responses had been obtained. The 39 input and response pairs were put in a random order and presented in that same order to all of the evaluators. 45 evaluators each read from 1 to 39 input and response pairs. Given each input and response pair, the evaluators were asked to rate the response by selecting either 1 (not a joke), 2 (almost a joke), 3 (a joke), or 4 (a very good joke).

I used that relatively simple 4-point scale because it directly measures the degree to which the evaluators perceive Witscript's responses to be jokes. If Witscript's responses are perceived to be jokes, then the system can be used to equip a conversational agent with humor, which would enhance the agent's performance (Dybala et al. 2008). And enhancing the performance of conversational agents such as chatbots was my main goal in creating Witscript.

Each input and response pair was evaluated by 15 workers and the mean of their ratings for the response in each pair was then calculated. Table 1 shows the first response that was output by the two systems for each input, together with the human response. All responses were transcribed exactly as they were output. The mean rating given by the evaluators for each response is also shown.

| Source | Input and responses | Rating |
|---|---|---|
| | **1. Input: A man pulled out a gun at a Popeyes restaurant when he heard they had no more chicken sandwiches.** | |
| DialoGPT | I think you mean a man pulled out a gun at a Popeye's restaurant when he heard they had no more chicken sandwiches. | 1.53 |
| Witscript | Popeyes gun? Um, not a Bubba Gun. | 2.33 |
| Human | He also objected to any new laws on bun control. | 3.00 |
| | **2. Input: America's national forests are being severely damaged by the illegal growing of marijuana.** | |
| DialoGPT | I think you dropped this : s | 1.73 |
| Witscript | America marijuana? Yah, so now he wants to be not in Canada cannabis. | 2.27 |
| Human | Which explains the new Forest Service mascot, Tokey the Bear. | 3.40 |
| | **3. Input: For Oktoberfest, Adidas is releasing sneakers that repel beer and vomit.** | |
| DialoGPT | I'm not sure if I want to see that. | 1.67 |
| Witscript | Adidas vomit? Ah, I like the Puma poop. | 2.87 |
| Human | And they're changing their slogan to "Just Spew It." | 3.27 |
| | **4. Input: Two men were arrested in the theft of a hundred thousand diapers.** | |
| DialoGPT | I'm sure they were arrested for the same thing. | 1.80 |
| Witscript | Theft diapers? Like, they stole stroller. | 2.00 |
| Human | They were charged with potty larceny. | 3.33 |
| | **5. Input: It's the twenty-fifth anniversary of the Blue Man Group.** | |
| DialoGPT | I was going to say the same thing. I was going to say it's the twenty proclamation of the blue man group. | 1.73 |
| Witscript | Blue anniversary? Heh, welcome to the Bluebilee. | 2.87 |
| Human | To celebrate, they're starting their own health plan: Blue Man Blue Cross Blue Shield. | 2.80 |
| | **6. Input: Thirteen TGI Fridays were selling cheap liquor in premium brand bottles.** | |
| DialoGPT | I was just there today and saw a bunch of them. | 1.33 |
| Witscript | Thirteen Fridays? Okay, but I prefer a nine night myself. | 2.07 |
| Human | The company claims it's a new promotion: TGI Fraudays. | 3.00 |
| | **7. Input: At a zoo, the first rhinoceros ever conceived by artificial insemination was born.** | |
| DialoGPT | I think you mean the first rhinoceros ever conceived by artificial insemination was born. | 1.60 |
| Witscript | Rhinoceros insemination? Mmm-hmm, welcome to the zoosemination. | 2.13 |
| Human | Usually rhinos are conceived using that dating site for large mammals, Ashley Mastodon. | 2.93 |

Table 1: Responses generated for each input, with the mean rating of each response on a scale from 1 (not a joke) to 4 (a very good joke).

|  | **8. Input: Dow Chemical is turning raw sewage into drinking water.** |  |
|---|---|---|
| DialoGPT | I think you mean sewage. | 1.60 |
| Witscript | Dow Chemical drinking? Yup, so not Dowing. | 2.13 |
| Human | To make it more appealing, they're calling it Pooland Spring. | 3.07 |
|  | **9. Input: A new study says drinking sugar-sweetened beverages increases the risk of death.** |  |
| DialoGPT | I think you mean to say that drinking sugar beverages increases the risk of death. | 1.27 |
| Witscript | Beverages risk? Huh, I have to focus on drinks risks. | 1.73 |
| Human | That's why they're renaming that soft drink Coca-Coma. | 2.87 |
|  | **10. Input: In one of its restrooms, the Guggenheim Museum is installing a solid gold toilet.** |  |
| DialoGPT | I think that's a new one. | 1.20 |
| Witscript | Gold toilet? Yep, to make a toilet pure. | 2.13 |
| Human | It's perfect for people who have just eaten 24 carrots. | 3.27 |
|  | **11. Input: A brewery apologized for a beer that shares the same name as a Hindu deity.** |  |
| DialoGPT | I think you mean a Hindu deity apologized for a beer that shares the same name as a brewery. | 2.00 |
| Witscript | Shares Hindu? Heh, no Sikh stock. | 2.87 |
| Human | No word yet on their plans for their other beer, Buddha-weiser. | 3.47 |
|  | **12. Input: A truck was stolen that contained half a million dollars' worth of handbags.** |  |
| DialoGPT | I'm sure it was a truck full of handbags. | 1.53 |
| Witscript | Truck handbags? Yep, of vanbags. | 2.13 |
| Human | The thief was arrested and the handbags were charged as accessories. | 2.87 |
|  | **13. Input: A man was caught during a home burglary wearing only lingerie.** |  |
| DialoGPT | I'm not sure if that's a good thing or a bad thing. | 1.87 |
| Witscript | Burglary lingerie? Yes, and the burglaries bikinis. | 2.13 |
| Human | He was charged with panty larceny. | 3.00 |

Table 1 (continued).

Table 2 compares the systems based on their ratings. The second column shows that Witscript's responses were rated, on average, about halfway between those of the DialoGPT baseline and the human, a professional comedy writer.

The last column of Table 2 shows the percentage of responses that the evaluators rated as "a joke" or "a very good joke" (i.e., the number of ratings of 3 and 4 divided by the total number of ratings). Witscript's responses were judged to be jokes 41.5% of the time, compared to only 17.9% of the time for the responses of DialoGPT. Witscript's response to Input #5 was actually rated higher than the human's response, which could be the first time ever that a machine defeated a human expert in a joke-writing challenge.

Some of the joke responses generated by Witscript don't make total sense. For example, the wordplay connecting Input #2 in Table 1 to Witscript's response is clear, but the logic isn't. Despite their occasional gaps in logic, fully 79.0% of Witscript's responses in Table 1 were rated by the evaluators as 2 (almost a joke) or higher. These impressive results may be partly due to the fact that each Witscript response has, at least, the form of a joke, complete with a punch line. Therefore, Witscript probably takes advantage of the "charity of interpretation" effect: evaluators may perceive that each of the well-formed linguistic containers offered up by Witscript is an even more meaningful joke than it actually is (Veale 2016). Still, if humans judge a Witscript response to be a joke, for whatever reasons, then the response is likely to function as a joke in a conversational context.

Those evaluation results, in connection with the research about joke-equipped chatbots cited in the Introduction, lead to the conclusion that users would perceive a chatbot equipped with a Witscript module to be more humanlike and likeable than one without it.

| System | Mean rating | % jokes (ratings of 3 or 4) |
|---|---|---|
| DialoGPT | 1.61 | 17.9% |
| Witscript | 2.28 | 41.5% |
| Human | 3.10 | 85.1% |

Table 2: Comparison of the systems based on their ratings

## Discussion

### Computational Creativity

I believe that the Witscript system demonstrates strong computational creativity instead of mere generation because its output exhibits three characteristics: novelty, value, and intentionality (Veale and Pérez y Pérez 2020; Ventura 2016).

The system's output has **novelty** because the contextually relevant joke that the system generates in response to a new input has almost certainly never been created before by it or by any other agent.

The system's output has **value**, as shown by the ratings given to its responses by human evaluators.

And the system produces that novel, valuable output with **intentionality** in several ways: It restricts its generation process by using domain knowledge about how a professionally-written joke is structured. It creates punch lines using pretrained word embeddings as a knowledge base for obtaining semantically related words. It completes jokes in an autonomous fashion by using a language model fine-tuned on an inspiring set consisting of professionally-written jokes. Finally, it employs a fitness function to rank the generated joke responses and intentionally filter out some that don't meet a preset threshold of value.

## Contributions

In addition to presenting an implementation of computational creativity that could make a chatbot more humanlike and likeable, this paper makes the following contributions:
1. It presents a novel system that can automatically improvise contextually relevant wordplay jokes in a conversation.
2. It presents a novel method for measuring the quality of wordplay and automatically identifying some jokes that aren't funny enough to be delivered.
3. It demonstrates how computational humor can be implemented with a hybrid of neural networks and symbolic AI, where the symbolic AI incorporates expert knowledge of comedy domain rules and algorithms.
4. It presents an extensible framework for generating original jokes based on an input sentence. That framework mirrors the Basic Joke-Writing Algorithm described in the Introduction.

## Future Work

The following work is needed to enable the Witscript system to execute the steps of the Basic Joke-Writing Algorithm more effectively and thereby output more sophisticated and funnier joke responses.

**Selecting Topic Keywords**  To consistently select the topic keywords that are the most potentially useful for punch line generation, the Witscript system needs to incorporate better natural language processing tools.

For example, with better named entity recognition, Witscript might have identified "TGI Fridays" in Input #6 in Table 1 as the name of a restaurant chain instead of the plural of a weekday. With better part-of-speech tagging of Input #11, Witscript might have identified "shares" as a verb instead of a noun related to the stock market.

Also needed is a method for selecting the pair of topic keywords most likely to capture the user's attention that is better than using vector-space distance. For example, a better method applied to Input #9 would have selected "beverages" and "death" instead of "beverages" and "risk."

**Generating Associations**  Research should be devoted to developing a more effective association engine to list associations, i.e., words and phrases related to a topic keyword. Sometimes Witscript will list words and word chunks whose relation to a topic keyword isn't obvious, which can lead it to assemble a weak, puzzling punch line.

For example, to generate its response to Input #10 in Table 1, Witscript came up with the obscure association "karat pure" for "gold"—probably as used in phrases like "18 karat pure"—and then substituted "toilet" for "karat" to create the weak punch line "toilet pure."

This limitation of Witscript's Word2Vec-based association engine is partly due to the fact that word embedding models such as Word2Vec define relatedness as the extent to which two words appear in similar contexts, which can be different from how closely associated two words are in the minds of humans (Cattle and Ma 2017). So future research might explore developing a more effective association engine that is a hybrid of a text-based external language approach and a word-association-based internal language approach (Deyne, Perfors, and Navarro 2016).

A more effective association engine to use in generating conversational jokes would also not be static, as Witscript's Word2Vec-based engine is, but instead would be regularly updated. Updating the association engine regularly would not only increase the size of its vocabulary but also ensure that its associations are capturing fresh topical relationships, such as current events and what most people think about them (Cattle and Ma 2017). To do a better job of simulating a witty human companion, Witscript needs an association engine that can accurately answer a question like "What do most people today think of when they hear the words [name of a celebrity]?"

**Generating an Angle**  Future work might improve the method that Witscript uses to generate the angles for its joke responses. The joke-tuned BERT model tends to generate angles that are fairly simple and not as specific to the input context as the human-written angles are.

But the BERT-generated angles do have the virtue of being short, which is good from a humor perspective: the shorter a joke is, the funnier it tends to be (Toplyn 2014). And the BERT model is capable of connecting a topic sentence to even the strangest punch line in a way that makes the system's output sound reasonably natural.

For example, in contributing to the responses for Inputs #5 and #7 in Table 1, the BERT model apparently decided that the unique portmanteaus "Bluebilee" and "zoosemination" could be names for the noteworthy occasions described by the inputs. So for an angle, the BERT model supplied a logical way to introduce a noteworthy occasion: "Welcome to the..." Only for Input #6 in Table 1 did the BERT model fail to generate useable text, which led to the system turning to a prewritten template to complete its response.

**Creating Punch Lines**  Currently the Witscript system creates punch lines that rely on wordplay. But as advances continue to be made in AI, this system could be extended to create punch lines by using Punch Line Makers that don't rely on wordplay (Toplyn 2014). Such Punch Line Makers include techniques that rely instead on common-sense knowledge to generate associations and to link them to create a punch line.

Consider this human-written example, adapted from a joke posted on the Twitter account @JoeToplyn:

Input: The U.S. is planning to buy 22 aging fighter jets from Switzerland.

Response: Yeah, the Swiss fighter jets have air-to-air missiles, smart bombs, a can opener, a nail file, and a toothpick.

If the Witscript system were equipped with common-sense knowledge, it might generate a joke response like the one in that human-written example by taking these steps:
1. Select as the topic keywords "fighter jets" and "Switzerland."
2. Determine that air-to-air missiles and smart bombs are parts of fighter jets.
3. Determine that a Swiss Army knife is related to Switzerland. Also determine that a can opener, a nail file, and a toothpick are parts of a Swiss Army knife.
4. Create a punch line that links the topic keywords "fighter jets" and "Switzerland" by generating this surprising juxtaposition of associations: "air-to-air missiles, smart bombs, a can opener, a nail file, and a toothpick."
5. Recall that the punch line is a list of parts of the entities specified by the topic keywords "fighter jets" and "Switzerland." Blend those topic keywords into the phrase "the Swiss fighter jets" so as to parallel the association "a Swiss Army knife." Use that "parts of" relationship to generate an angle by appending to "the Swiss fighter jets" the verb "have."
6. Concatenate a filler word, the angle, and the punch line to get the final joke response: "Yeah, the Swiss fighter jets have air-to-air missiles, smart bombs, a can opener, a nail file, and a toothpick."

**Selecting the Best Joke**  The current implementation of the Witscript system determines which joke response to output by selecting the punch line candidate that incorporates the best-scoring wordplay. But future implementations of the system could employ a more comprehensive funniness score to determine which joke response candidate to output. The funniness score of a joke response candidate could comprise several feature scores, which would be weighted and combined to yield the funniness score.

Those feature scores could comprise the wordplay score of the joke's punch line and also an interest score measuring the degree to which the two topic keywords used to create the punch line are related to each other. The less related that the topic keywords are to each other, the more attention-getting they may be when appearing together in the topic, the better the interest score, and the better the funniness score (Toplyn 2020b).

The feature scores could also include a clarity score measuring the degree to which the two topic keywords are related to their respective associations that were linked to create the joke's punch line. The more closely related each topic keyword is to its association that was used to create the joke's punch line, the more understandable the joke, the better the clarity score, and the better the funniness score (Toplyn 2020b).

## Conclusion

In this paper we have introduced Witscript, a novel joke generating system that can improvise conversational joke responses that depend on wordplay. The Witscript system, because it embodies the Basic Joke-Writing Algorithm, also provides a road map to generating more sophisticated joke responses that depend on common-sense knowledge.

But because the Witscript system in its current implementation seems to regularly generate acceptable joke responses, it could even now be integrated as a humor module into an open-domain chatbot (Sjobergh and Araki 2009). The proprietary software is available for license.

People often use chatbots to fulfill a desire for entertainment or socializing (Brandtzæg and Følstad 2017). So an open-domain chatbot that uses Witscript to occasionally ad-lib a pretty good joke might potentially animate an artificial, but likeable, companion for lonely humans.